\documentclass{article}
\usepackage{spconf,amsmath,graphicx}

\title{Integrating Categorical Features in End-to-End ASR}
\name{Rongqing Huang}
\address{Apple \\ \small \tt{huangr@apple.com}}
\begin{document}
%
\maketitle
\begin{abstract}
All-neural, end-to-end ASR systems gained rapid interest from the speech recognition community. Such systems convert speech input to text units using a single trainable neural network model. E2E models require large amounts of paired speech text data that is expensive to obtain. The amount of data available varies across different languages and dialects. It is critical to make use of all these data so that both low resource languages and high resource languages can be improved.  When we want to deploy an ASR system for a new application domain, the amount of domain specific training data is very limited. To be able to leverage data from existing domains is important for ASR accuracy in the new domain.  In this paper, we treat all these aspects as categorical information in an ASR system, and propose a simple yet effective way to integrate categorical features into E2E model. We perform detailed analysis on various training strategies, and find that building a joint model that includes categorical features can be more accurate than multiple independently trained models.

\end{abstract}
\section{Introduction}
\label{sec:intro}

In recent years, all-neural, end-to-end (E2E) models that directly convert speech into text through a sequence model have become popular. Instead of separately optimized components, an E2E model is a single trainable neural network. It removes the HMM assumptions and enables end-to-end optimization. Architectures like Connectionist Temporal Classification (CTC) \cite{ref:graves14}, attention based sequence models such as Listen, Attend and Spell (LAS) \cite{ref:las15}, transformer \cite{ref:vaswani17, ref:dong18}, and Recurrent Neural Network Transducer (RNN-T) \cite{ref:graves12} have obtained impressive results and sometimes surpass conventional systems \cite{ref:chiu18, ref:jinyuli20, ref:boli21}. 

Though E2E models are theoretically strong, they require large amounts of paired speech text data which is usually expensive to obtain. 
The amount of paired speech text data also varies across different languages and locales. Locale-specific ASR systems can be less accurate in under-represented locales simply because there is insufficient transcribed data. When developing an ASR system for a new application domain, there is typically little domain specific transcribed data. Thus the model quality suffers. In this study, we treat the different languages, dialects, domains as categorical information to the ASR system. In this paper, we propose a simple method to inject the categorical information into the E2E model, which allows the pooling as much data as possible to train the E2E model. Not only are we able to train better models for the less represented languages and domains, but the performance on the primary language and domain is also improved. We also find this joint model performs better than an individually fine-tuned model. This is desirable since we have less models to maintain and deploy.

The multilingual E2E model has been studied previously.
In \cite{ref:zhou18}, language information is added as a token in the beginning or the end of the output sequence. In 
 \cite{ref:li18dialect, ref:toshniwal18, ref:kannan19}, the language information is inserted into the model directly either as an embedding or one-hot vector. In \cite{ref:pratap20}, separate decoders are used for different groups of languages.   The main difference in our work is we expand the idea into multiple categorical features, not only including language and dialect, but domain and potentially other types of categorical features like device types and contexts. We propose a simple method to combine multiple categorical features and compare different ways to inject the resulting combination into the model. Another contribution is we perform extensive experiments and analysis of strategies in multilingual and multi-domain E2E training. We use an LAS model in our study, although the method should be applicable to other types of E2E models like RNN-T or transformer.

\section{Base LAS Model}
\label{sec:background}

Given a sequence of speech frames $\textbf{x}=\left\{x_1,...,x_L\right\}$ with length $L$, and output token sequence $\textbf{y}=\left\{y_1,...,y_U\right\}$ with length $U$, the E2E model computes the probability
\begin{equation}
P(\textbf{y}|\textbf{x})=\prod_{i=1}^{U}P(y_i|\textbf{x},y_1,y_2,...,y_{i-1})
\end{equation}
An encoder converts $\textbf{x}$ to intermediate outputs $\textbf{h}=\left\{h_1,...,h_T\right\}$ through a Recurrent Neural Network (RNN), typically an LSTM. One important aspect for ASR is there are many more speech frames than the number of output tokens in $\textbf{y}$, usually there is a reduction factor $N$, thus $T=\frac{L}{N}$ is the length of encoder output.
\begin{equation}
h_j = \mathrm{EncoderRNN}(x_j, h_{j-1})
\label{eq:enc}
\end{equation}
The decoder is also an RNN. It takes encoder outputs $\textbf{h}$ (a.k.a. memory), and previous output token $y_{i-1}$, generates the current decoder state $s_i$:
\begin{equation}
s_i = \mathrm{DecoderRNN}(y_{i-1}, s_{i-1}, c_i)
\label{eq:dec}
\end{equation}
which is then passed through a generation network, typically a feedforward network with softmax output, to produce the next output token $y_i$:
\begin{equation}
y_i=\mathrm{Generate}(s_i, c_i)
\end{equation}
$c_i$ is the context vector at decoder step $i$ that summarizes information from the encoder:
\begin{equation}
c_i=\sum_{j=1}^{T}\beta_{i,j}h_j
\end{equation}
where $\beta_{i,j}$ is attention weight at output step $i$ on $j$-th encoder output.

\section{Injection of Categorical Features}
\label{sec:method}
The categorical information like language, dialect, domain is sentence based. One way to inject a categorical feature $f_k$ into the model is by using an embedding:
\begin{equation}
e_k=E_k(f_k)
\end{equation}
for the $k$-th categorical feature. We can combine $M$ categorical features into a single feature vector:
\begin{equation}
e=\sum_{k=1}^{M}V_{k}e_{k} + b_k
\label{eq:catf_wt}
\end{equation}
where $V_k$ is a weight matrix, and $b_k$ is a bias vector for $k$-th categorical feature. All embeddings and weights are trained as part of the whole E2E model.
\subsection{Encoder}
There are various places the categorical feature vector $e$ can be inserted into the model \cite{ref:li18dialect, ref:toshniwal18, ref:kannan19}. We can append it to the input feature in the encoder:
\begin{equation}
\textbf{x} = \left\{[x_1, e], [x_2, e], ..., [x_L, e]\right\}
\end{equation}
\subsection{Decoder}
The categorical feature vector $e$ can also be inserted into the decoder by concatenating it with the context vector $c_i$. Specifically, Eq. \ref{eq:dec} becomes
\begin{equation}
s_i = \mathrm{DecoderRNN}(y_{i-1}, s_{i-1}, [c_i, e])
\end{equation}

\subsection{Encoder and decoder}
The categorical feature vector $e$ can be inserted into both the encoder and decoder. For increased flexibility, we apply different weight matrices $V_{k,enc}$, $V_{k,dec}$, and bias vectors $b_{k,enc}$, $b_{k,dec}$ in Eq. \ref{eq:catf_wt} for encoder and decoder respectively. In the experiment section, we will present results of each method.

\section{Experiments}
\label{sec:exp}

\subsection{Data}
We use Chinese data in all our experiments. The data includes following four language/dialects: Mandarin, Cantonese, Taiwanese, and Shanghainese. \footnote{We use the terms language and dialect loosely and interchangeably here.} For Mandarin and Shanghainese, Simplified Chinese characters are used; for Cantonese and Taiwanese, Traditional Chinese characters are used. Each dialect has various number of domains: assistant (ast), message (msg), search (srch), map, in-home smart speaker (home), in-car smart speaker (car). Mandarin has all the domains, while Shanghainese has only message domain. The amount of training data available for each dialect and domain is listed in Table \ref{tab:train_data}. Each domain in a dialect has a test set. The amount of data for each test set ranges from 10 to 40 hours. We only perform domain dependent modeling experiments in Mandarin because other dialects do not have enough domain specific data.  

\begin{table}[th]
\caption{Amount of training data across dialects and Mandarin domains}
\label{tab:train_data}
\centering
\begin{tabular}{|l|c|c|}
\hline
\textbf{Dialect} & \textbf{Hours} & \textbf{text} \\
\hline
Mandarin & 18k & Simplified Chinese \\
\quad \textit{assistant (ast)} & \textit{7k} & \\
\quad \textit{message (msg)} & \textit{4k} & \\
\quad \textit{home} & \textit{2k} & \\
\quad \textit{search (srch)} & \textit{500} & \\
\quad \textit{car} & \textit{400} & \\
\hline
Cantonese & 7k & Traditional Chinese \\
\hline
Taiwanese & 3k & Traditional Chinese \\
\hline
Shanghainese & 100 & Simplified Chinese \\
\hline
\end{tabular}
\end{table}

\subsection{The E2E model} 
The baseline E2E model is LAS. The encoder is a 5-layer bi-directional LSTM, 1200 cells in each direction, with 600-dim projections.
The 80-dim log mel-filter bank feature is the input to the encoder. A 4-head, 1200-dim attention is used. The decoder uses a 2-layer unidirectional LSTM, each has 800 cells. All the training texts are merged and processed with 19k BPE tokens derived from the training data. The model has 178m parameters. 

The dialect and domain features are modeled with separate 80-dim embedding tables, the transformed dimension is 20 and 160 for injecting into encoder and decoder respectively. The value is chosen so the total amount of increased parameters is around 200K regardless of where the categorical features are injected. This amounts to only 0.1\% parameter increase over the baseline model. 

The E2E model is trained with block-momentum SGD \cite{chen2016scalable}, 32 V100 GPUs, 10000 frames per minibatch per node, L2 normalization on gradient, SpecAugment \cite{ref:park19} on the encoder input, with label smoothing \cite{ref:szegedy16} weight 0.05 and schedule sampling \cite{ref:bengio15} weight 0.1. The cross entropy training is followed by one epoch of Minimum Word Error Rate training (MWER, \cite{prabhavalkar2018, shen2016}). The weight on the cross entropy loss during MWER is 0.05. The initial learning rate is 0.025 and decayed by factor 0.8 when no improvement observed on the validation set. 
We use beam size 8 during beam search and length penalty \cite{ref:wu16} 0.1.

\subsection{Different experiment setups}
We will present results from a few independently trained, jointly trained, and fine-tuned models. Table \ref{tab:setup} explains the different setups in the study. For fine-tuning after joint training, the setup denoted with suffix $d$ (dialect) or $t$ (targeted domain) means fine-tuned to the dialect and fine-tuned to the targeted domain in the dialect, respectively.

\begin{table}[th]
\caption{Different experiment setups in the study}
\label{tab:setup}
\centering
\begin{tabular}{|l|l|}
\hline
\textbf{Setup} & \textbf{Description} \\
\hline
S0 & dialect independently trained \\
\hline
S1 & dialect and domain independently trained \\
\hline
S2 & jointly trained \\
\hline
S3 & jointly trained, categorical feature in encoder \\
\hline
S4 & jointly trained, categorical feature in decoder \\
\hline
S5 & jointly trained, cat. feature in encoder \& decoder \\
\hline
S*d & fine-tune a jointly trained model to dialect \\
\hline
S*t & fine-tune a jointly trained model to targeted domain \\
\hline
\end{tabular}
\end{table}

\subsection{Joint trained models}
\label{sec:catf_result}
We could simply pool all the data together and train a single model. This is a baseline (setup S2) for comparing the effectiveness of injecting categorical features into the model. Please note that all the results for setup S2-S5 are obtained using a single shared model for all dialects and domains. The jointly trained model results are presented in Table \ref{tab:catf}.

\begin{table}[t]
\caption{The CER (Character Error Rate) of independently and jointly trained models on different dialects and domains.}
\label{tab:catf}
\centering
\begin{tabular}{|l|c|c|c|c|c|c|}
\hline
\textbf{Setup}  & \textbf{Ast} & \textbf{Car} &\textbf{Home} & \textbf{Map} &\textbf{Msg} &\textbf{Srch} \\
\hline
\multicolumn{7}{l}{\textbf{Mandarin}}\\
\hline
S0 & 5.44 & 5.91   & 2.92  & 7.35  & 6.11  & 10.36  \\
\hline
S1 & 5.87 & 14.85 & 4.25 & 8.24 & 7.22 & 10.16 \\
\hline
S2 & 5.99 & 6.25 & 3.19  & 7.61 & 6.80 & 11.0   \\
\hline
S3 & 5.27 & 5.66  & 2.57 & \textbf{6.88} & 6.04 & 9.78  \\
\hline
S4 & 5.34 & 5.75  & \textbf{2.47} & 7.07 & \textbf{5.77} & 9.95   \\
\hline
S5 & \textbf{5.23} & \textbf{5.62}  & 2.57 & 6.90 & 5.91 & \textbf{9.68}  \\
\hline
\multicolumn{7}{l}{\textbf{Cantonese}}\\
\hline
S0 & 6.16 & 5.07  & 2.96 & 9.47 & 5.21 & 10.62  \\
\hline
S2 & 6.66 & 5.68 & 3.04  & 10.48 & 5.64 & 11.60  \\
\hline
S3 & \textbf{5.60} & 5.00 & 2.66 & 9.23 & 5.14 & 9.20   \\
\hline
S4 & 5.72 & 4.99  & \textbf{2.63} & 9.18 & \textbf{5.01} & 9.65   \\
\hline
S5 & 5.63 & \textbf{4.85}  & \textbf{2.63} & \textbf{8.68} & 5.19 & \textbf{9.19}  \\
\hline
\multicolumn{7}{l}{\textbf{Taiwanese}} \\
\hline
S0 & 5.89 & 5.64  & 4.79 & 7.28 & 5.91 & 12.97  \\
\hline
S2 & 10.12 & 9.38 & 7.43 & 10.11 & 6.86 & 17.38  \\
\hline
S3 & \textbf{4.15} & 4.68 & \textbf{3.44}  & \textbf{5.56} & 4.34 & \textbf{9.26}   \\
\hline
S4 & 4.44 & 5.04 & 3.47  & 6.09 & 4.52 & 10.08   \\
\hline
S5 & 4.35 & \textbf{4.57} & 3.47  & 5.58 & \textbf{4.31} & 9.35  \\
\hline
\multicolumn{7}{l}{\textbf{Shanghainese}} \\
\hline
S0 & - & - & - & - & 53.11 & -  \\
\hline
S2 & - & - & - & - & 20.03 & - \\
\hline
S3 & - & - & - & - & 15.83 & -   \\
\hline
S4 & - & - & - & - & 16.48 & -  \\
\hline
S5 & - & - & - & - & \textbf{14.63} & -  \\
\hline
\end{tabular}
\end{table}

\begin{table}[t]
\caption{The CER of fine-tuned models on different dialects and domains, along with non-fine-tuned results.}
\label{tab:fine_tune}
\centering
\begin{tabular}{|l|c|c|c|c|c|c|}
\hline
\textbf{Setup}  & \textbf{Ast} & \textbf{Car} &\textbf{Home} & \textbf{Map} &\textbf{Msg} &\textbf{Srch} \\
\hline
\multicolumn{7}{l}{\textbf{Mandarin}}\\
\hline
S2 & 5.99 & 6.25 & 3.19  & 7.61 & 6.80 & 11.0 \\
\hline
S2d & 5.61 & 6.14 & 2.92 & 7.38 & 6.45 & 10.62  \\
\hline
S2t & 5.51 & 6.07  & 2.80 & 7.21 & 6.17 & 10.16 \\
\hline
S5 & 5.23 & 5.62  & 2.57 & 6.90 & 5.91 & 9.68 \\
\hline
S5d & \textbf{5.16} & 5.59  & \textbf{2.47} & 6.96 & 5.90 & 9.70 \\
\hline
S5t & \textbf{5.16}  & \textbf{5.55}  & 2.48 & \textbf{6.86} & \textbf{5.85}  & \textbf{9.63} \\
\hline
\multicolumn{7}{l}{\textbf{Cantonese}}\\
\hline
S2 & 6.66 & 5.68 & 3.04  & 10.48 & 5.64 & 11.60  \\
\hline
S2d & 5.70 & 4.89 & 2.75  & 8.95 & 4.76 & 9.65  \\
\hline
S5 & 5.63 & 4.85  & \textbf{2.63} & 8.68 & 5.19 & 9.19 \\
\hline
S5d & \textbf{5.59}  & \textbf{4.77}  & 2.64  & \textbf{8.53} & \textbf{4.63}  & \textbf{9.05}  \\
\hline
\multicolumn{7}{l}{\textbf{Taiwanese}}\\
\hline
S2 & 10.12 & 9.38 & 7.43 & 10.11 & 6.86 & 17.38  \\
\hline
S2d & 4.28 & 4.66 & 3.61  & 5.56 & 4.38 & 9.65  \\
\hline
S5 & 4.35 & 4.57 & 3.47  & 5.58 & 4.31 & 9.35 \\
\hline
S5d & \textbf{4.20}  & \textbf{4.28}   & \textbf{3.40}  & \textbf{5.25}  & \textbf{4.19}  & \textbf{8.88}  \\
\hline
\multicolumn{7}{l}{\textbf{Shanghainese}}\\
\hline
S2 & - & - & - & - & 20.03 & - \\
\hline
S2d & - & - & -  & - & 14.60 & -  \\
\hline
S5 & - & -  & - & - & 14.63 & -  \\
\hline
S5d & - & -  & - & - & \textbf{14.15} & - \\
\hline
\end{tabular}
\end{table}

\subsection{Fine-tune model to dialect and domain}
After the jointly trained model is finished, we can fine-tune this model to each dialect and domain. The fine-tuning process is straightforward: beginning with the best jointly trained model, we continue training with dialect or domain specific data until convergence. In the end we will have one model for each dialect or domain.
From Table \ref{tab:catf}, we see that a single model injected with categorical features perform well across dialects and domains (setup S3-S5). A natural followup question is, what if we can fine-tune the base model and deploy a dialect/domain specific model, will a single model still be the best? Overall, we perform this fine-tuning process for setup S2 and S5 in Table \ref{tab:fine_tune}.

\begin{table}[th]
\caption{Average CER for each setup and dialect}
\label{tab:avg}
\centering
\begin{tabular}{|l|c|c|c|c|}
\hline
\textbf{Setup} & \textbf{Mandarin} & \textbf{Cantonese} & \textbf{Taiwanese} & \textbf{Shanghainese} \\
\hline
S0 & 6.35 & 6.58 & 7.08 & 53.11 \\
\hline
S1 & 8.43 & - & - & - \\
\hline
\hline
S2 & 6.81 & 7.18 & 10.21  & 20.03 \\
\hline
S3 & 6.03 & 6.14 & 5.24  & 15.83 \\
\hline
S4 & 6.06 & 6.20 & 5.61 & 16.48 \\
\hline
S5 & 5.98 & 6.03 & 5.27 & 14.63 \\
\hline
\hline
S2d & 6.52 & 6.12 & 5.36 & 14.60 \\
\hline
S2t & 6.32 & - & - & - \\
\hline
S5d & 5.96 & \textbf{5.87} & \textbf{5.03} & \textbf{14.15} \\
\hline
S5t & \textbf{5.92} & - & - & - \\
\hline
\end{tabular}
\end{table}

\section{Analysis}
In this section, we present detailed analysis on experimental results. For easier comparison, we calculate the unweighted mean CER across all tasks in Table \ref{tab:avg}.

Simply pooling all data together (S2) does not perform better than individually trained dialect dependent models (S0), except for extremely low resource dialects like Shanghainese (compare S0 and S2 in Table \ref{tab:avg} and Table \ref{tab:catf}).

Fine-tuning the S2 model with dialect dependent data greatly helps accuracy. S2d models are better than S0 models on lower resource dialects now (Table \ref{tab:avg}). However, it still cannot match the dialect dependent model on the high resource dialect (compare S0 and S2d for Mandarin in Table \ref{tab:avg}). Perhaps because the modeling capacity is redirected from the main dialect to other dialects. 

Domain dependent models (S1) do not work better than dialect dependent models (S0) except on the search task (compare Mandarin S1 and S0 in Table \ref{tab:catf}). This is probably because there are similar utterances in various domains. Sharing data across the domains makes the model more robust. However, if a model sees data from all domains and is later fine-tuned to a specific domain, there is still an accuracy benefit.  The domain adapted model has a small advantage over the dialect adapted model (compare Mandarin S2t and S2d in Table \ref{tab:fine_tune}). The average relative CERR (Character Error Rate Reduction) is 3\%. 

Incorporating categorical features into the E2E model gives significant accuracy boost (compare S3, S4, and S5 with S2 in Table \ref{tab:avg}). The relative CERR is larger on low resource dialects (comparing S5 with S2, 48\% and 27\% on Taiwanese and Shanghainese respectively), and smaller on high resource dialects (12\% and 16\% on Mandarin and Cantonese respectively). Similar to \cite{ref:kannan19}, we also find inserting the feature into encoder (S3) is better than inserting into decoder (S4). However, we also observe inserting into both the encoder and decoder (S5) gives small but consistent gains over inserting the feature into the encoder only (S3). Note that in all cases, the model size is only increased by 0.1\% over the baseline E2E model used in S0, S1, and S2.

It is also important to note that the S3/S4/S5 setups use one model for all the dialects. This has a huge development and deployment advantage over the dialect specific model setups S0 and S2d. In our study, S5 also outperforms S0 and S2d by 28\% and 3\% respectively (averaged over all dialects).

We see further accuracy gains from fine-tuning the S5 model to individual dialects. Comparing S5d with S5 in Table \ref{tab:avg}, we find no gain on Mandarin, but still 3-4\% relative CERR for the other dialects. Fine-tuning on the individual domains gives insignificant additional gain (compare Mandarin S5t with S5d in Table \ref{tab:avg}).

Overall, if only one model is used for all dialects studied here, S5 achieves 12-48\% relative CERR over S2. If each dialect has its own model, S5d achieves 3-9\% relative CERR over S2d.   

\section{Conclusion}
Dialect, language, domain etc categorical information is important knowledge for improving ASR. In this study, we propose a method to incorporate multiple categorical features into an E2E model. We find that injecting the combined feature into both encoder and decoder performs the best. We train a single model with categorical features for 4 Chinese dialects and 6 domains. This model (S5) performs better than dialect independently trained models (S0) and dialect adapted models (S2d) by 28\% and 3\% respectively. Further gains are achieved by fine-tuning to language/domain. The model S5d achieves 3-9\% relative CERR over baseline dialect adapted model (S2d). In the future, we plan to extend this categorical feature into more languages and other types of categorical information. 

\section{Acknowledgments}
We thank Donald McAllaster, Ernest Pusateri, John Bridle, Russ Webb, and Woojay Jeon for helpful feedback.
\vfill\pagebreak

\bibliographystyle{IEEEbib}
\bibliography{refs}

\end{document}